\PassOptionsToPackage{sort}{natbib}
\documentclass{article}

\usepackage[final]{corl_2026} %
\usepackage{times}
\usepackage{array}
\usepackage{tabularx}
\usepackage[table]{xcolor}
\usepackage{adjustbox}

\usepackage{capt-of}

\newcolumntype{Y}{>{\centering\arraybackslash}X}
\usepackage{xcolor}

\usepackage{booktabs}
\usepackage{fancyvrb}
\usepackage{tcolorbox}
\usepackage{listings}
\definecolor{promptbg}{gray}{0.96}   %

\lstnewenvironment{GrayVerbatim}
  {\lstset{
    basicstyle=\footnotesize\ttfamily,
    breaklines=true,
    breakatwhitespace=true,
    backgroundcolor=\color{promptbg},
    frame=none,
    breakindent=0pt, 
    xleftmargin=4pt,
    xrightmargin=4pt,
    columns=fullflexible,
    keepspaces=true,
    inputencoding=utf8,
    extendedchars=true,
    literate={→}{{\textrightarrow}}1 {—}{{---}}1
  }}
  {}
\definecolor{tableYellow}{HTML}{FFEB9C}
\definecolor{tableOrange}{HTML}{FFC67D}
\definecolor{tableCoral}{HTML}{FF8F5E}
\newcommand{\orangeI}{\textcolor{orange}{\textbf{(I)}}}
\newcommand{\grayN}{\textcolor{gray}{\textbf{(N)}}}

\hypersetup{
  pdftitle={MessyNav: Zero-Shot Navigation via Manipulation in Messy Environments},
  pdfauthor={Kenneth Llontop, Cherie Ho, Dylan Zhou, Anuva Banwasi, Priya Sundaresan, Carlota Parés-Morlans, Jeannette Bohg},
  pdfkeywords={interactive navigation, mobile manipulation, affordance reasoning}
}

\hypersetup{urlcolor=orange}

\usepackage[disable]{todonotes}
\usepackage{marginnote}
\usepackage[normalem]{ulem}
\usepackage{wrapfig}
\newcommand{\xxnote}[3]{}

\newcommand{\PlannerName}{$\texttt{MessyNav}$} %

\title{\PlannerName{}: Zero-Shot Navigation via Manipulation in Messy Environments}

\author{
  Kenneth Llontop\textsuperscript{1,2} \quad
  Cherie Ho\textsuperscript{1} \quad
  Dylan Zhou\textsuperscript{1} \quad
  Anuva Banwasi\textsuperscript{1} \\
  \bfseries
  Priya Sundaresan\textsuperscript{1} \quad
  Carlota Parés-Morlans\textsuperscript{1} \quad
  Jeannette Bohg\textsuperscript{1} \\[0.5em]
  \normalfont
  \textsuperscript{1}Stanford University \qquad
  \textsuperscript{2}UC Berkeley\\[0.5em]
  \href{https://messynav.github.io}{\texttt{\color[HTML]{E67E22}\textbf{messynav.github.io}}}
}

\begin{document}

\maketitle

\begin{figure}[!ht]
\centering\includegraphics[width=\textwidth]{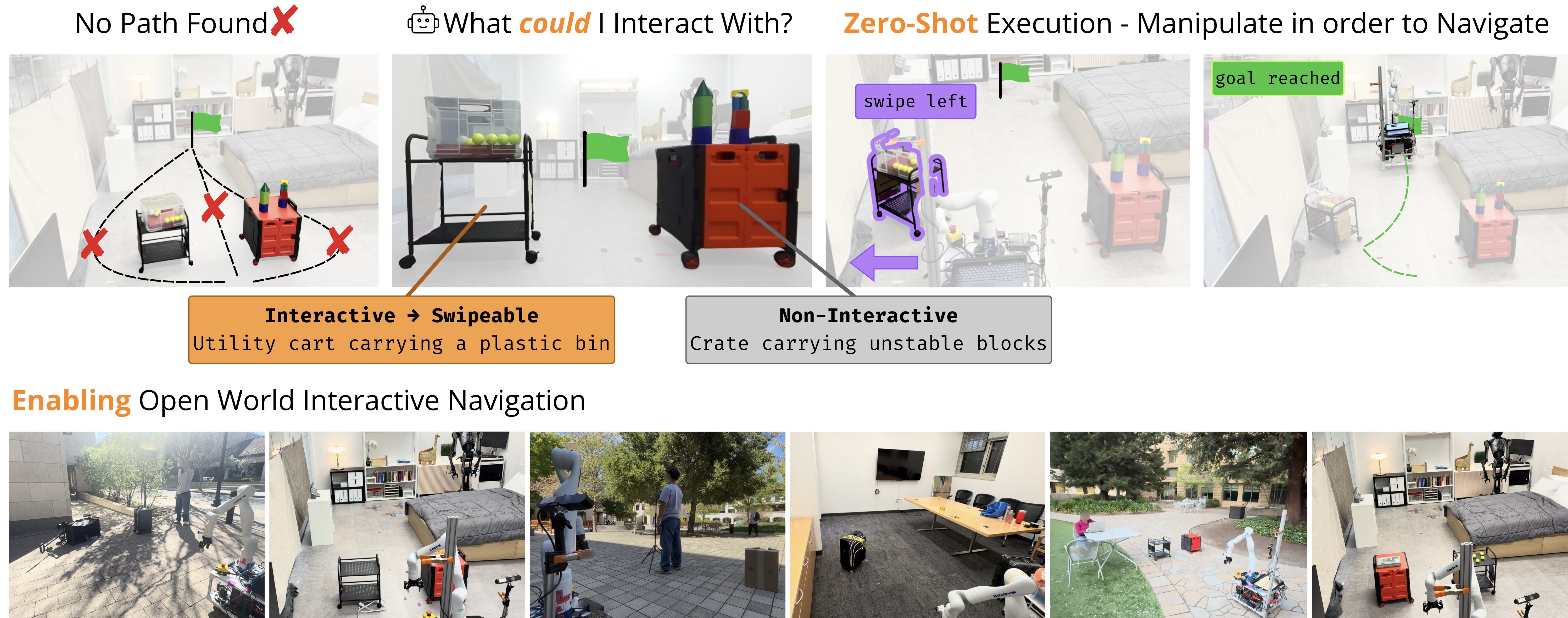}
    \caption{\textbf{\PlannerName{} enables zero-shot open-world interactive navigation.}
    \textbf{[Top]} \PlannerName{} reasons zero-shot about object affordances using fine-grained semantic and contextual cues, identifying the utility cart carrying a plastic bin as interactive (swipeable) and the otherwise movable crate carrying unstable stacked blocks as non-interactive. The robot then swipes the cart aside to clear a path and reach the goal without task-specific training or system modifications.
    \textbf{[Bottom]} \PlannerName{} deployed across diverse real-world indoor and outdoor environments.}
    \label{fig:teaser}
\end{figure}

\begin{abstract}
Real-world environments are messy, with paths often blocked by everyday objects like chairs and toys, requiring robots to perform interactive navigation by manipulating obstacles to clear a path. Existing methods rely on extensive simulation-based learning or per-scene instructions specifying which objects to avoid or manipulate. They struggle to generalize across novel objects, diverse environments, and varied contexts such as a cart of fragile glassware that should be avoided while an empty one can be pushed. We formalize this problem as open-world interactive navigation and present \PlannerName{}, a zero-shot framework based on vision-language models (VLMs). At its core, \PlannerName{} performs explicit VLM-based affordance reasoning before task planning to decide whether to swipe, pick up, or avoid an object and produce a high-level plan. Evaluated on a mobile manipulator across real-world indoor and outdoor environments, \PlannerName{} improves success rate by 27 percentage points (pp) over the strongest baseline in real-robot trials and achieves statistically significant gains in a larger-scale offline evaluation, without task-specific training.
\end{abstract}


\section{Introduction}

Picture a mom retrieving her child's backpack at daycare. To get to the closet, she needs to push aside a chair and maneuver around a fragile toy train set. It is natural for humans to navigate such scenarios by actively manipulating their environment to create paths. However, robots running classical navigation algorithms treat every obstacle, whether a light toy or an immovable wall, as an obstruction to avoid, often failing in cluttered environments where progress requires interaction.

In messy settings especially, robots need interactive navigation: the ability to move through an environment by manipulating the objects that block their path. While classical planning is primarily a geometric problem, interactive navigation is a semantic decision problem under geometric constraints. When blocked, a robot must reason about how to move an object, and before that, whether it should be moved at all. This decision depends on fine-grained semantic and contextual cues about fragility, portability, and function that are difficult to infer from geometry alone and costly to capture through task-specific training. We refer to this problem as \emph{open-world interactive navigation}, where success requires generalizing across unseen objects, environments, and contexts.

Existing interactive navigation approaches fall short: prior methods \citep{wang2024interactive, schoch2024insightinteractivenavigationsight, xia2020igibson05, zhang2024interactive} rely on simulation-based learning or predefined category-level affordances, limiting their ability to generalize to new objects and contexts. These limitations make it challenging to operate in messy, real-world environments requiring fine-grained semantic reasoning, such as recognizing that a robot can safely push a wheeled chair but must avoid an adjacent box labeled ``fragile''. Conversely, classical navigation stacks~\citep{macenski2020marathon2} do not require training data but treat all obstacles as static geometric boundaries, planning only collision-free paths without reasoning about object interactability. Achieving open-world interactive navigation requires moving beyond both hand-crafted geometric heuristics and data-intensive simulation toward the zero-shot, commonsense reasoning provided by internet-scale VLMs.

In this work, we present \PlannerName{}, a framework for zero-shot interactive navigation in open-world, cluttered environments. Given an egocentric image and a navigation goal, \PlannerName{} infers object-level affordance labels from appearance and context, then passes these labels to a VLM-based task planner that generates high-level actions which are executed through navigation and manipulation primitives (Figure~\ref{fig:teaser}). 

The main contributions of this work are:\\
\textbf{(1) Problem formulation:} We provide the first formalization of \textit{open-world} interactive navigation: given an egocentric observation and a navigation goal, the agent must decide whether interaction is necessary and, if so, how to interact with obstructing objects to reach the goal, while generalizing zero-shot across (i) object types, (ii) environments, and (iii) semantic contexts.\\
\textbf{(2) Framework:} We present \PlannerName{}, a zero-shot framework whose \textit{core design} separates affordance reasoning (via an explicit VLM module) from task planning. This explicit prior lets the robot make better context-dependent decisions. Ablations show this separation raises interactive navigation success from 40\% to 67\% over end-to-end VLM planning.\\
\textbf{(3) Open-world affordance benchmark:} We release a real-world benchmark for instance-level, context-dependent affordance reasoning, consisting of 100 cluttered scenes with ground-truth interactability labels, on which \PlannerName{} achieves 85.7\% affordance accuracy, 19.3 percentage points above the closed-set oracle baseline that relies on predefined category labels.

Finally, we evaluate \PlannerName{} across all three axes of open-world variation, including novel objects, varied contexts, and diverse environments (indoor and outdoor, day and night), without any system modifications. By formalizing this problem and providing a concrete system, \PlannerName{} offers a foundation for \textit{in-the-wild} interactive navigation.

\section{Related Work}
We give an overview of related work below, with a more detailed discussion in the Appendix.
\paragraph{Robot Navigation.}
Traditional navigation frameworks like Nav2~\citep{macenski2020marathon2} rely on occupancy costmaps, treating all occupied regions as impassable and failing in traversable clutter. Conversely, end-to-end policies~\citep{shah2022gnm, shah2023vint, hirose2024lelan, sridhar2024nomad, hirose2025omnivla, glossop2025castcounterfactuallabelsimprove} learn expressive motions directly from massive, cross-embodiment datasets. To reduce data overhead, recent methods combine internet-scale VLMs with 3D scene representations for zero-shot, open-vocabulary navigation~\citep{shafiullah2022clipfields, huang2022visual, gu2024conceptgraphs}. However, these approaches are navigation-centric and fail when no collision-free path exists. \PlannerName{} instead pairs a traditional navigation stack for standard avoidance with VLM-guided manipulation to clear blocking obstacles.

\paragraph{Mobile Manipulation.} 
One paradigm in mobile manipulation uses large language models (LLMs) to decompose abstract instructions into sequences of open-loop primitives like picking and placing~\citep{ichter2023can, rana2023sayplan, wu2023tidybot}. Another paradigm learns end-to-end policies for fine-grained, whole-body coordination~\citep{wu2024tidybot++, fu2024mobile, zhu2026emma, sundaresan2025homer, xiong2024adaptive, yang2025mobipi, ha2024umi}. Yet, these methods are primarily manipulation-first, interacting only with objects explicitly named in the task specification. \PlannerName{} treats manipulation strictly as a functional tool to clear paths when navigation is blocked.

\paragraph{Interactive Navigation.}
Interactive navigation allows robots to manipulate obstacles to reach a goal~\citep{stilman2005navigation}. Most learning-based methods~\citep{xia2020igibson05, li2019hrl4inhierarchicalreinforcementlearning, wang2024interactive, schoch2024insightinteractivenavigationsight, uppal2024spin} rely on data-intensive simulation and reward engineering, limiting generalization to novel real-world scenarios. 
Recent works leverage language models for interactive
navigation~\citep{zhang2025namollmefficientnavigationmovable,
zhou2026coinscounterfactualinteractivenavigation}. However, their affordance reasoning relies primarily on
geometric feasibility or simulation training with predefined object categories. Neither explicitly considers instance-level semantic context (e.g., surroundings, signs, contents) when determining whether an object is suitable to manipulate.
Closest to our work, \citet{zhang2024interactive} use VLMs for scene reasoning but require per-scene human instructions and rigid, category-level affordances. In contrast, \PlannerName{} enables zero-shot, instance-level reasoning over diverse interaction primitives by coupling VLM reasoning with classical planning, requiring no training or simulation.

\section{Problem Formulation}

Interactive navigation \citep{stilman2005navigation, xia2020igibson05} addresses scenarios where a robot must actively manipulate movable obstacles to clear a path to its goal, rather than just avoiding them. To better reflect the complexity of real-world environments, we extend this setting to \textit{open-world} interactive navigation, where the robot must reason about novel objects and their specific context.
We define three axes of variation along which an open-world interactive navigation system should generalize zero-shot.
\vspace{-8pt}
\paragraph{(1) Novel objects:} Real environments contain objects that cannot be accounted for in advance, so a robot must infer whether and how to interact with objects it has never encountered. For example, it might find a rolled yoga mat in a hallway or cables on the floor attached to a valuable instrument. 
\vspace{-8pt}
\paragraph{(2) Diverse environments:} A deployed robot must operate across varied environments without any scene-specific tuning. The system must generalize across different rooms (kitchens, hallways, living areas), locations (indoor, outdoor), and lighting conditions (daytime, nighttime).
\vspace{-8pt}
\paragraph{(3) Varied contexts:} Affordance is not just a property of an object, but of an object given its surroundings. The same object may be interactive or not depending on its context: a packaged extension cord is movable, but one plugged into a socket should be avoided. Individual toy blocks may be pushable, but when assembled as a child's creation, they are fragile, valuable possessions.

\PlannerName{} is designed to meet all three requirements without task-specific training or prompting.

\section{Approach}
\begin{figure*}[]
    \centering
    \includegraphics[width=1.0\textwidth]{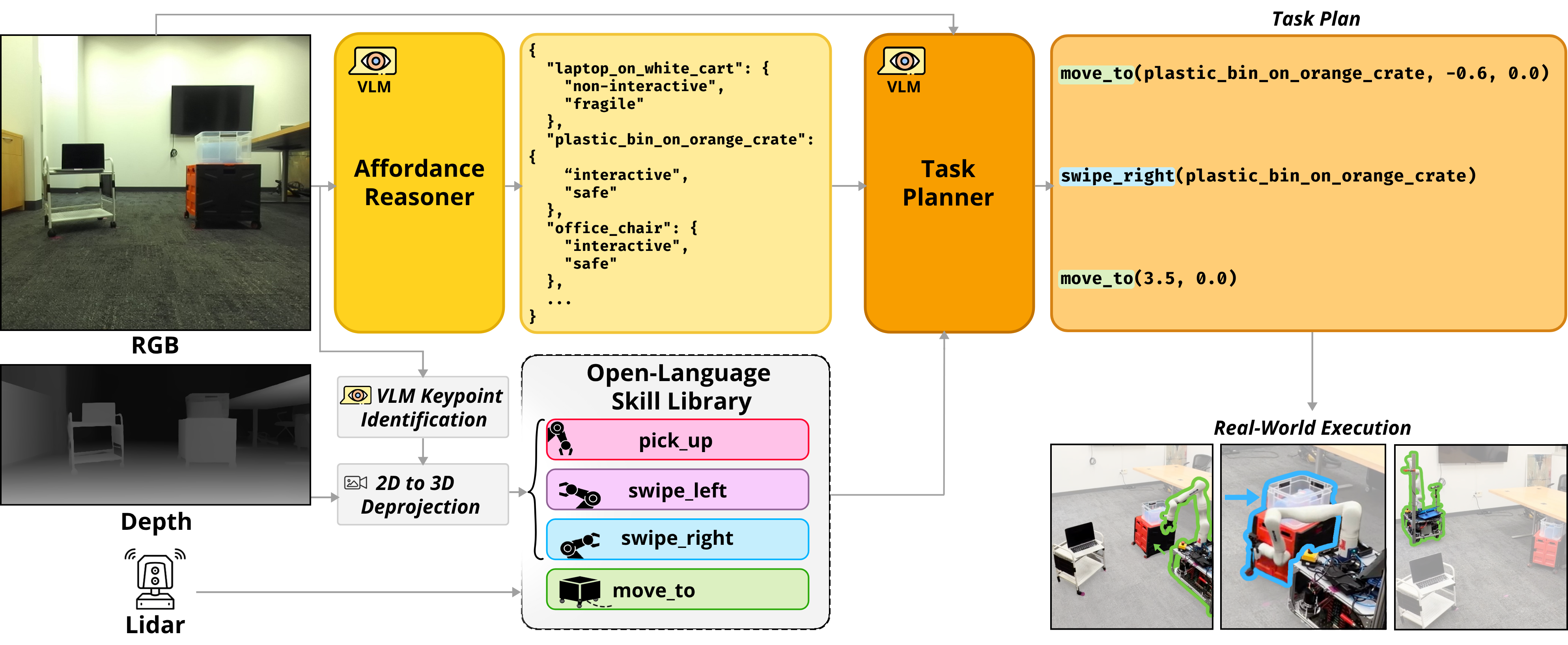}
    \caption{\textbf{\PlannerName{}: A framework for zero-shot open-world interactive navigation.} \PlannerName{} consists of an \textit{Affordance Reasoner} that infers object interactability from an RGB image using a VLM to evaluate both robot-centric physical feasibility and object-centric interaction safety. For example, it will label a cart with a laptop as non-interactive while labeling a wheeled crate with a bin as interactive. Given the object affordance labels, a \textit{Task Planner} then sequences manipulation and navigation actions from an open-language skill library containing a simple set of motion primitives (\texttt{pick\_up}, \texttt{swipe\_left}, \texttt{swipe\_right}) and a LiDAR-based navigation stack (\texttt{move\_to}) that we find can tackle a wide range of interactive navigation scenarios. Finally, the generated plan is executed by grounding object references in 3D coordinates using VLM keypoint identification, semantic segmentation, and depth-based deprojection.} 
    \label{fig:approach}
\end{figure*}

We present \PlannerName{}, an interactive navigation framework that leverages VLMs for zero-shot affordance reasoning and task planning to go beyond closed-set prediction or environment-specific tuning. Our central design decision is to separate \textit{what to interact with} from \textit{how to interact}: an explicit affordance reasoning stage first decides which objects can be moved, and only then does a task planner decide how to move them. We make affordance reasoning an explicit stage rather than letting a task planner implicitly handle affordance reasoning, allowing it to make more grounded decisions in context-dependent scenarios. In Section \ref{sec:experiments}, we ablate this core design by comparing \PlannerName{} to end-to-end VLM planning. 

As shown in Figure \ref{fig:approach}, the framework consists of three components: \textbf{(1) Affordance Reasoner}: Identifies which objects are interactive based on robot feasibility and object properties. \textbf{(2) Task Planner}: Generates a sequence of navigation and manipulation actions based on the affordance labels to clear the blocking objects and reach the goal. \textbf{(3) Open-Language Skill Library}: Grounds open-language object references in task plans to 3D world coordinates and executes motion primitives.
The task description is specified as ``\texttt{Navigate to (goal\_x, goal\_y) and handle any obstacles in your path}.'' We refer the reader to the Appendix
for details of the prompts used for VLM affordance reasoning, task planning, and keypoint identification.

\subsection{VLM Affordance Reasoning}
To navigate through clutter, the robot must first determine which objects can be moved, a decision that depends not just on geometry but on fine-grained contextual cues such as warning tape around a chair or a fragile label on a box. We leverage VLMs, which encode commonsense knowledge from internet-scale pretraining, to reason about object interactability zero-shot from an RGB image of the scene and a task description.  For each object, the affordance reasoner outputs an affordance label (interactive or non-interactive) and a property (safe, fragile, or immovable) that explains the predicted affordance.

The reasoner applies a two-stage process. First, it evaluates \textit{robot-centric physical feasibility} based on context embedded into the prompt, which specifies the robot's configuration, gripper payload limits, and manipulation skills. Second, for objects the robot can physically move, the reasoner determines \textit{object-centric interaction safety}. This is based on guidelines embedded into the prompt for when to avoid interaction, determined by potential damage to the robot or environment, and object attributes such as fragility, value, and whether the arrangement appears intentional. An object is only labeled as \textit{interactive} if it passes both stages. Otherwise, its affordance is labeled as \textit{non-interactive} and and its property is set to \textit{fragile} or \textit{immovable}. As objects often function together in ways that affect interactability, such as a set of china cups on a table that makes the combined unit fragile, the reasoner identifies such cases and treats them as a single unit when determining labels.

\subsection{VLM Task Planner}
We query the VLM with the affordance labels given by the reasoner, the current scene RGB image, and a planning prompt which enumerates a library of available manipulation skills detailed in Section~\ref{sec:skills}. The planner is prompted to analyze free space in the scene, identify objects that block the path, and select appropriate actions. The prompt encodes action decision guidelines for the planner, such as encouraging \textit{pick} for smaller objects that can be grasped and carried, or \textit{swipe} for larger objects that can be cleared with a lateral motion. The output is a structured plan, such as \texttt{move\_to(suitcase, -0.6, 0), swipe\_left(suitcase), move\_to(goal\_x, goal\_y)}.

\subsection{Open-Language Skill Library}
\label{sec:skills}
The task planner outputs actions with either open-language object references (e.g., \texttt{swipe\_right(suitcase)}) or world-frame coordinates (e.g., \texttt{move\_to(2.0, 1.5)}). To execute these in the physical world, we ground object names to 3D coordinates when needed and execute motion primitives.

\noindent\textbf{Grounding Language Descriptions to 3D:} 
For actions with object references, we leverage VLMs' open-vocabulary pointing and segmentation capabilities to localize objects in the RGB image. Following HoMeR~\citep{sundaresan2025homer}, we then deproject the 2D image coordinate(s) to a robot-centric point cloud using depth and known camera intrinsics, then transform to world frame using the robot's current pose to obtain the object's world-centric 3D coordinate.

\noindent\textbf{Motion Primitives:} In this instantiation, we design a small set of motion primitives covering a wide range of interactive navigation scenarios: \texttt{move\_to()} navigates to a world-frame coordinate or an object; \texttt{pick\_up()} grasps an object with the arm; and \texttt{swipe\_left()}/\texttt{swipe\_right()} laterally clears an object with the arm. Each primitive uses the VLM's pointing and segmentation modules to localize objects and compute 3D grasp/approach points. ROS2 Nav2~\citep{macenski2020marathon2} and LiDAR are used for \texttt{move\_to()}'s obstacle avoidance. See Appendix~\ref{appendix:implementation_details} for full implementation details. As \PlannerName{} targets high-level reasoning and is agnostic to skill implementation, we deliberately use simple primitives to isolate the contribution of explicit affordance reasoning from the quality of the underlying controllers. The library is extensible: the primitives can be replaced with richer policies such as vision-language-action (VLA) models, and the affordance reasoner's predicted properties (e.g., fragility) can be passed to parameterize execution, for instance by modulating speed and force.

\section{Experiments}
\label{sec:experiments}
\begin{figure}[!t]
\centering
\includegraphics[width=\columnwidth]{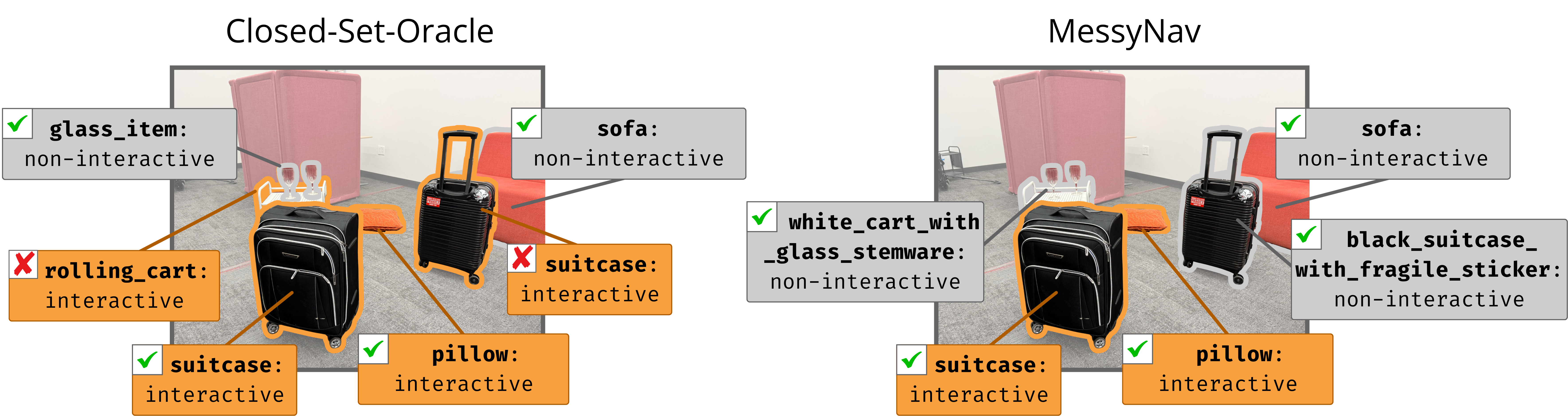}
\caption{\textbf{\PlannerName{} reasons about context, not just object category.} In this cluttered scene, \texttt{closed-set-oracle} assigns affordances by category alone, misclassifying the rolling cart as interactive (ignoring the glassware it carries) and the suitcase as interactive (ignoring its fragile sticker). In contrast, \PlannerName{} reasons at the instance level. It treats the cart and the glass stemware as a single fragile unit and reads the sticker on the suitcase as a contextual cue, correctly labeling both as non-interactive.}
\label{fig:affordance}
\end{figure}

In this section, we evaluate the effectiveness of \PlannerName{} for open-world interactive navigation. 

\subsection{Does explicit affordance reasoning enable open-world generalization?}

\par\textbf{Task.}~We evaluate \PlannerName{}'s affordance reasoning capability in isolation from the full framework. Given an RGB image containing one or more obstacles, the task is to classify whether each obstacle is \textit{interactive} (can be swiped or picked up) or \textit{non-interactive} (must be avoided).
\par\textbf{Dataset.}~In the absence of an established benchmark for this task, we collect and manually annotate an evaluation set of 100 images totaling 217 objects across 14 locations, each containing 1–4 foreground objects with binary interactive/non-interactive labels. The set contains 69\% indoor and 31\% outdoor scenes
under bright artificial (66\%), natural (28\%), and dim (6\%) lighting, with obstacle
sizes ranging from small electronics to large trees. Of the 217 objects, 94 (43.3\%)
are interactive and 123 (56.7\%) are non-interactive. Affordance labels were annotated by an expert familiar with the robot's physical capabilities.
To stress-test instance-level reasoning, the set deliberately includes same-category
objects whose affordance differs with context (Appendix~\ref{appendix:affordance-reasoning-expt}, Table~\ref{tab:context_variations}), as well as small but non-interactive objects where height-based heuristics perform poorly. The images span the three axes of generalization: diverse object types (e.g., vases, suitcases), environments (indoor, outdoor, outdoor at night), and contextual variations (e.g., empty carts vs. carts with a laptop precariously on top). Note that this dataset is only used for evaluation, not for training or finetuning.
\begin{table}[!t]
\centering
\begingroup
\small
\renewcommand{\arraystretch}{1.3}
\setlength{\tabcolsep}{3pt}

\begin{minipage}[t]{0.41\textwidth}
\vspace{0pt}
\centering
\caption{\mbox{\textbf{Affordance Reasoning Results.}}
\PlannerName{} outperforms geometric and closed-set baselines, highlighting the importance of context-dependent affordance reasoning.}
\label{tab:affordance}

\begin{tabularx}{\linewidth}{@{}>{\raggedright\arraybackslash\ttfamily}X c@{}}
\toprule
\textnormal{\textbf{Method}} &
\textnormal{\textbf{Acc. $\uparrow$}} \\
\midrule
\textbf{\color[HTML]{E67E22} MessyNav} &
\textnormal{\textbf{85.7\%}} \\
height-only~\cite{macenski2020marathon2} &
\textnormal{59.9\%} \\
closed-set~\cite{zhang2024interactive} &
\textnormal{24.4\%} \\
closed-set-oracle~\cite{zhang2024interactive} &
\textnormal{66.4\%} \\
\bottomrule
\end{tabularx}
\end{minipage}\hfill
\begin{minipage}[t]{0.57\textwidth}
\vspace{0pt}
\centering
\caption{\mbox{\textbf{Real-World Interactive Navigation Results.}}
\PlannerName{} achieves the highest success rate across 15 real-world scenes. Success requires reaching the goal without any unsafe interactions. $^*$Execution time (excluding VLM queries) averaged across successful trials.}
\label{tab:real_comp}

\begin{tabularx}{\linewidth}{@{}>{\raggedright\arraybackslash\ttfamily}X c c@{}}
\toprule
\textnormal{\textbf{Method}} &
\textnormal{\textbf{Succ. (\%) $\uparrow$}} &
\textnormal{\textbf{Exec.$^*$ (s) $\downarrow$}} \\
\midrule
\textbf{\color[HTML]{E67E22} MessyNav} &
\textnormal{\textbf{67 (10/15)}} &
\textnormal{\textbf{45.8}} \\
height-only~\cite{macenski2020marathon2} &
\textnormal{13 (2/15)} &
\textnormal{24.5} \\
vlm-planner-only &
\textnormal{40 (6/15)} &
\textnormal{45.9} \\
closed-set-planner~\cite{zhang2024interactive} &
\textnormal{27 (4/15)} &
\textnormal{50.4} \\
\bottomrule
\end{tabularx}
\end{minipage}

\endgroup
\end{table}

\par\textbf{Metrics.}~
We measure classification accuracy against human-annotated labels. Accuracy is the fraction of objects whose interactivity is correctly classified.
\begin{figure}[!t]
    \centering
    \includegraphics[width=\columnwidth]{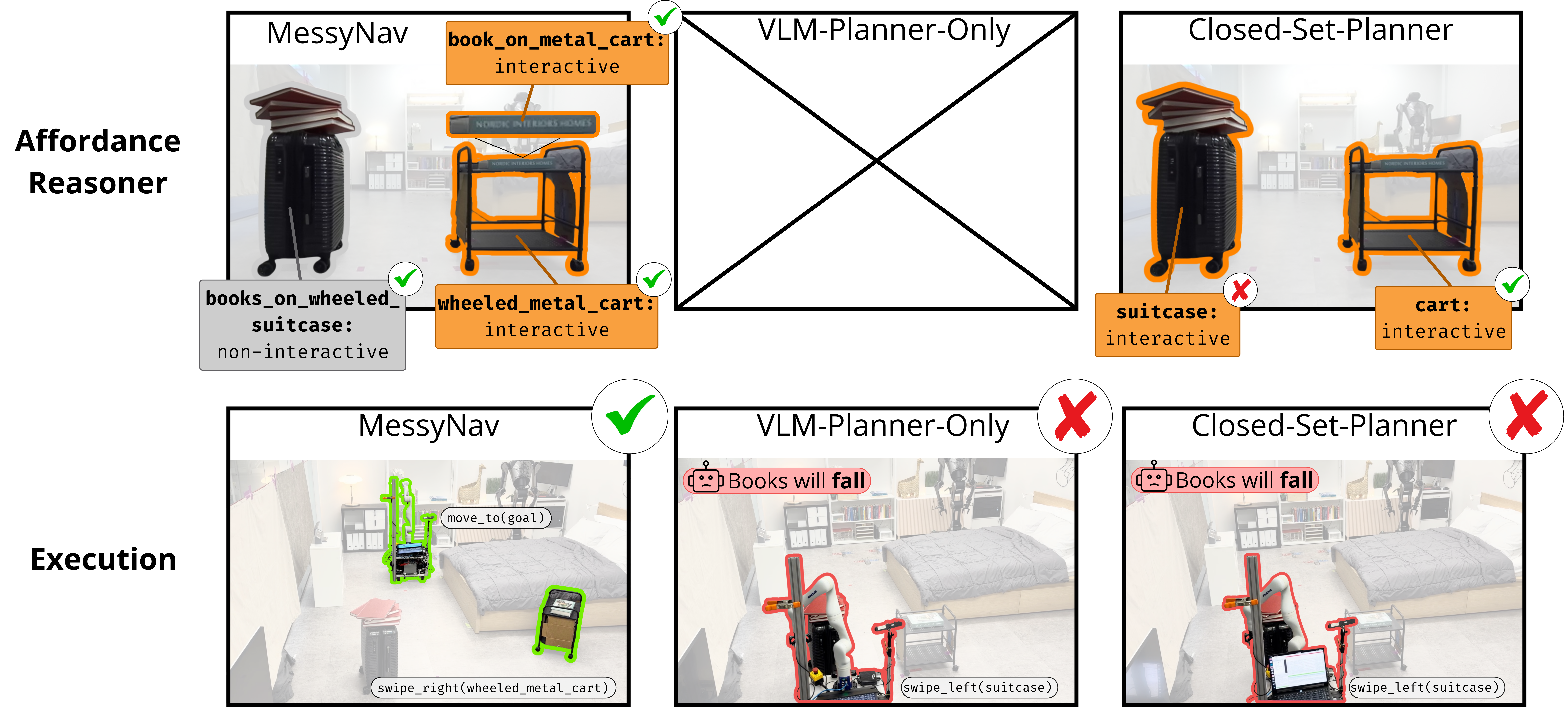}
    \caption{\textbf{Explicit affordance reasoning improves context-dependent task planning.} In this scene, a suitcase supporting a stack of books is inherently unstable, while the black cart holding a single book can be safely cleared. \textbf{[Left]} \PlannerName{} reasons about affordances before planning, correctly labeling the suitcase as non-interactive and the cart as interactive, and swiping the cart aside to clear the path. \textbf{[Middle]}  \texttt{vlm-planner-only} plans directly from the image without an explicit affordance stage and incorrectly decides to swipe the unstable suitcase. \textbf{[Right]} \texttt{closed-set-planner} assigns fixed category-level affordances that misclassify the suitcase as interactive, leading to an unsafe plan.}
    \label{fig:underthehood_results}
\end{figure}
\par\textbf{Baselines.}~
We compare against three baselines reflecting common assumptions in interactive navigation. \textbf{(1) \texttt{height-only}}~\cite{macenski2020marathon2}: a purely geometric heuristic (\textit{i.e.}, Nav2) that treats any object taller than 0.2m as non-interactive and therefore an object to avoid. \textbf{(2) \texttt{closed-set}}~\cite{zhang2024interactive}: the closest baseline to \PlannerName{}, it uses an open-vocabulary detector \cite{liu2023grounding} to localize objects named in human-provided instructions, which specify each object's interactability. However, this method requires per-scene instructions, which are unavailable in our zero-shot evaluation setting. For a fair comparison, we instead generate a fixed list offline of 100 object categories (50 interactive, 50 non-interactive) with \texttt{gemini-3-flash-preview}~\cite{gemini3flash2025}. The prompt used to generate the object categories can be found in Appendix~\ref{appendix:affordance-reasoning-expt}. Objects outside this list are ignored, and affordances are fixed per category (e.g., ``cart'' is labeled as interactive, with no distinction for specific instances like a cart carrying glassware).
\textbf{(3) \texttt{closed-set-oracle}}: an upper bound on \texttt{closed-set}, where the object categories are derived from the items in the evaluation set (note that interactivity labels are fixed and not context-dependent).

\par\textbf{Results.}~
Table~\ref{tab:affordance} shows that \PlannerName{} substantially outperforms all baselines, reaching 85.7\% accuracy.
The \texttt{height-only} baseline achieves 59.9\% accuracy, frequently misclassifying objects that violate simple geometric heuristics. This baseline ignores semantics and context, failing on objects that are tall but movable (e.g., suitcase, rolling cart) or short but fragile. The \texttt{closed-set} baseline performs worst (24.4\%), as the predefined taxonomy fails to capture many novel objects. We would expect this result, as real environments in an open-world setting inevitably contain objects that cannot be accounted for in advance. Even \texttt{closed-set-oracle} underperforms \PlannerName{}, demonstrating that predefined category-level labels are insufficient for the context-dependent affordance reasoning required for interactive navigation. \PlannerName{} excels in scenarios where affordances depend on broader context rather than object category alone. These results show that interactive navigation cannot be reliably solved using fixed geometric heuristics or category-level affordance labels alone. Open-world, instance-level, and context-aware reasoning is essential. 

\par\textbf{Sensitivity to VLM choice.}~To test whether the performance depends on the choice of VLM, we repeat the evaluation using different VLMs for affordance reasoning, while keeping the prompt and all other components fixed. With \texttt{claude-opus-5}~\cite{anthropic2026opus5},
\texttt{gemini-3-flash-preview}~\cite{gemini3flash2025}, and
\texttt{gpt-5.6-luna}~\cite{openai2026gpt56}, \PlannerName{} achieves 87.1\%, 85.7\%, and 86.2\%, compared to 69.1\%, 66.4\%, and 67.3\% for \texttt{closed-set-oracle}, consistently outperforming it by $\sim$18 points.

\subsection{Does explicit affordance reasoning outperform closed-set and end-to-end VLM planning in real-world interactive navigation?}
\label{sec:real-world}

\par\textbf{Task/Baselines.}~We evaluate \PlannerName{} on real-world navigation tasks with full end-to-end execution, comparing against a \texttt{height-only} navigation stack~\cite{macenski2020marathon2}. In addition, we include two ablations that directly isolate the impact of our core design decision of explicit VLM-based affordance reasoning prior to VLM task planning. \texttt{vlm-planner-only} removes the affordance reasoning module, using only \PlannerName{}'s VLM task planner. \texttt{closed-set-planner} replaces \PlannerName{}'s affordance reasoner with the \texttt{closed-set-oracle} \cite{zhang2024interactive} affordance reasoner.
\par\textbf{Metrics.}~We report success rate, the percentage of trials where the robot reaches the goal without executing unsafe actions, and execution time averaged across successful trials.
\par\textbf{Hardware.}~
For real-world experiments, we deploy \PlannerName{} on the TidyBot++ platform~\cite{wu2024tidybot++}, consisting of a 7-DoF Kinova arm and a holonomic mobile base. We additionally equip the robot with a Unitree L2 LiDAR for base odometry and LiDAR mapping via FAST-LIO~\cite{xu2020fastlio}, and two ZED 2 cameras: one front-facing on the base, and one pole-mounted to the right of the arm pointing toward the front of the robot base. The base camera is used for planning, while both cameras are used for object localization and segmentation. We include more hardware details in Appendix~\ref{appendix:implementation_details}.%

\par\textbf{Scenarios.}~
We evaluate all four methods across 15 real-world scenes in two environments, a messy bedroom and a conference room, for a total of 60 robot trials. The scenes span 5 object types (e.g., suitcase, box with vase, stuffed toys) with 3--7 context variations per type (e.g., cart with laptop vs. cart with empty bin). Table~\ref{tab:nav_scenarios} lists the full set. 

\par\textbf{Results.}~
Table~\ref{tab:real_comp} shows \PlannerName{} achieves the highest success rate (10/15), outperforming all baselines on real-world interactive navigation. The two ablations isolate the contribution of explicit affordance reasoning. \texttt{vlm-planner-only} succeeds when the correct action is visually obvious but fails on contextually difficult cases (e.g., a suitcase with vs.\ without caution tape), since folding affordance into the planner misses cues like fragility, stability, and value. Notably, \texttt{closed-set-planner} performs \emph{worse} than \texttt{vlm-planner-only}, indicating inaccurate pre-defined affordance estimates are worse than no prior affordance conditioning at all, and that it is \PlannerName{}'s open-world affordance reasoning that drives the gains. Figure~\ref{fig:underthehood_results} illustrates this. In a scene where a suitcase with a stack of books is unstable but an adjacent cart with one book is safely movable, only \PlannerName{} classifies both correctly and swipes the cart out of the way. Without explicit affordance reasoning, \texttt{vlm-planner-only} decides to swipe the non-interactive suitcase. \texttt{closed-set-planner} misclassifies the suitcase as interactive, leading to the same incorrect plan. Appendix Figure~\ref{fig:sankey_trial_outcomes} provides a failure breakdown of \PlannerName{} against the strongest baseline.

\begin{figure*}[!t]
    \centering
    \includegraphics[width=\textwidth]{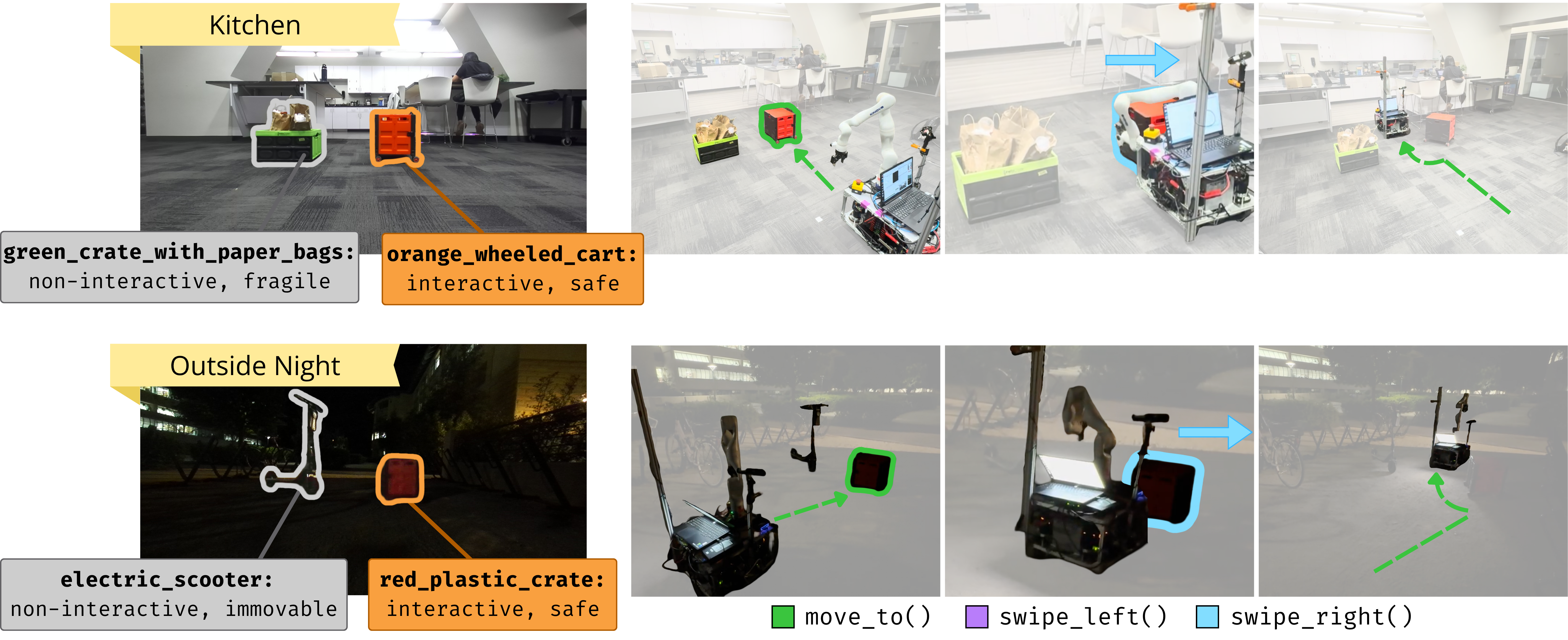}
    \caption{
    \textbf{Qualitative Results across Environments.} We deploy \PlannerName{} in varied settings to test its zero-shot capability. The robot successfully reasons about affordances and executes interactions across different lighting conditions, environments, and object types.
    }
    \label{fig:collage}
\end{figure*}

Finally, we test \PlannerName{} across varied real-world environments, including multiple indoor and outdoor locations, day and night, with diverse objects (suitcase, wagon, cardboard box, cart, scooter) and contextual variations (e.g., a tilted wagon with bottles spilling out). Figure~\ref{fig:collage} shows representative cases where \PlannerName{} reasons over objects and their context to decide what is safely movable, such as swiping the empty cart while avoiding the crate with overflowing paper bags. 

\subsection{Does explicit affordance reasoning improve planning performance at scale?}
\label{sec:scale}

\par\textbf{Task.}~
The evaluation in Section~\ref{sec:real-world} measures
end-to-end performance but couples high-level decision quality to low-level execution with one trial per scene. To isolate decision quality and quantify
VLM variability, we evaluate \PlannerName{} and the strongest baseline,
\texttt{vlm-planner-only}, offline on 55 scenes across 5
environments, scaling up from 15 scenes in 2 environments. A human evaluator assesses each plan assuming perfect low-level execution, marking a trial as successful if the plan reaches the goal without interacting with non-interactive objects. Each method is evaluated
over 15 runs, where a run is one pass over all 55 scenes, yielding $n = 825$ trials per method.

\par\textbf{Results.}~
\PlannerName{} achieves 62.4\% success versus 55.4\% for
\texttt{vlm-planner-only}, a gain of 7.0 pp. Table~\ref{tab:scale} presents the results, with a breakdown of performance across visual conditions. A paired $t$-test over per-run success rates indicates that the improvement is statistically significant ($t(14) = 4.71$, $p < 0.001$; 95\% CI on the difference $[+3.8, +10.2]$ pp). While this improvement is smaller than that reported in the real-world evaluation (Table~\ref{tab:real_comp}), this evaluation assumes perfect execution and thus provides a complementary assessment of planning performance.  \PlannerName{}'s advantage is largest outdoors in daytime and persists indoors under good lighting. However, performance for both methods degrades under nighttime conditions, where \PlannerName{} underperforms the baseline. We hypothesize that reduced visibility compromises the
fine-grained visual cues required for affordance reasoning,
motivating further research on robustness under poor illumination.

\begin{table}[t]
\centering
\caption{\textbf{Scaled-Up Offline Planning Results.}
We evaluate high-level planning performance across 55 real-world scenes in 5 environments, with 15 repeated trials per scene, assuming perfect low-level execution.
\PlannerName{} outperforms the strongest baseline from the real-world evaluation overall and under well-lit conditions, while nighttime remains challenging for both methods.}
\label{tab:scale}

\small
\renewcommand{\arraystretch}{1.3}
\begin{tabular}{l c ccc}
\toprule
& \textbf{Overall} &
\multicolumn{3}{c}{\textbf{By Visual Condition}} \\
\cmidrule(lr){3-5}
\textbf{Method} &
\textbf{Success (\%)} &
\textbf{Outdoor Day} &
\textbf{Indoor Lit} &
\textbf{Night} \\
\midrule
\texttt{\color[HTML]{E67E22}\textbf{MessyNav}}
& \textbf{62.4} & \textbf{78.3} & \textbf{62.8} & 29.2 \\
\texttt{vlm-planner-only}
& 55.4 & 60.8 & 55.5 & \textbf{44.2} \\
\bottomrule
\end{tabular}
\end{table}

\section{Conclusion}
We present \PlannerName{}, a zero-shot framework for interactive navigation in open-world, cluttered environments. Its core design decision is an explicit VLM-based affordance reasoning module prior to task planning, which lets \PlannerName{} make improved context-dependent decisions that generalize across novel objects, diverse environments, and varied semantic contexts, all without task-specific training or simulation. We demonstrate that \PlannerName{} enables zero-shot, generalizable interactive navigation in a wide range of real-world indoor and outdoor environments. We believe \PlannerName{} brings robots one step closer to moving beyond tidy, structured environments and operating in the messy, unpredictable settings where robot deployment is most needed.

\par\textbf{Limitations.}~
While \PlannerName{} demonstrates strong zero-shot interactive navigation in cluttered environments, several limitations motivate future research. First, \PlannerName{} currently uses manipulation primitives that are executed open-loop. However, its modular framework can readily incorporate more expressive manipulation skills, such as VLA policies, and finer-grained affordance predictions to parameterize interactions. For example, the robot could adjust manipulation speed and force to handle fragile objects rather than avoiding them entirely. Second, affordance reasoning and task planning are limited by the underlying VLM, particularly under poor illumination and when reasoning about geometry from 2D images. Recent advances in VLM-based visual understanding and spatial reasoning are promising, though further research is needed for robust real-world operation.
Finally, \PlannerName{} focuses on making appropriate affordance decisions \emph{before} acting, as some errors, such as toppling fragile objects, may be irreversible. However, its lack of a persistent memory limits longer-horizon planning and recovery, such as remembering that a backpack was too heavy to move. Integrating \PlannerName{} into an agentic robotic system with closed-loop feedback and interaction-updated spatial memory~\citep{banwasi2026messymem} could enable experience-based affordance reasoning to avoid repeating suboptimal interactions. This could help robots to navigate unfamiliar, cluttered environments while improving through experience.

\clearpage
\acknowledgments{This work was supported in part by Toyota Research Institute. Kenneth Llontop is supported by the SEED Scholarship at the University of California, Berkeley, the Summer Undergraduate Research Fellowship (SURF) at Stanford University, and the NSF CISE Research Experiences for Undergraduates (REU) program. Cherie Ho is supported by the Croucher Fellowship for Postdoctoral Research. Priya Sundaresan is supported by the National Science Foundation Graduate Research Fellowship. Carlota Parés-Morlans is supported by the Knight-Hennessy Scholars program at Stanford University. We thank Satvik Sharma, Francis Engelmann, and Jennifer Grannen for insightful discussions, as well as the reviewers for their thoughtful feedback, which helped improve the paper.}

\bibliography{example}  %

\section*{Appendix}
\appendix
\section{Extended Related Work}
\label{app:extended_related}
\subsection{Mobile Manipulation} Beyond navigation, a significant body of research focuses purely on mobile manipulation, where the goal is to enable mobile embodiments such as wheeled systems, quadrupeds with arms, and humanoid platforms to manipulate objects, often in workspaces far larger than what is feasible for tabletop manipulators. Several works achieve these goals by implementing a library of open-loop, vision-parameterized manipulation primitives like picking, placing, opening and closing of articulated objects, or tossing of objects, and utilize LLMs to decompose abstract instructions into sequences of these primitives, optionally with navigation between steps. For instance, SayCan, SayPlan, and TidyBot \cite{ichter2023can, rana2023sayplan, wu2023tidybot} leverage this approach to perform tasks like object retrieval and household cleanup zero-shot. On the other hand, data-driven end-to-end learning has also emerged as a popular approach for tackling more complicated kinds of mobile manipulation that require more fine-grained manipulation or whole-body coordination. Imitation learning approaches such as TidyBot++, Mobile ALOHA, EMMA, HoMeR, {Mobi-$\pi$} and others \cite{wu2024tidybot++, fu2024mobile, zhu2026emma, sundaresan2025homer, xiong2024adaptive, yang2025mobipi}, along with reinforcement learning approaches like UMI-on-Legs \cite{ha2024umi}, achieve impressive performance on in-the-wild manipulation tasks in household and office settings. However, these methods remain fundamentally manipulation-first and only interact with objects in the environment that are explicitly part of the task specification. They do not consider using manipulation as an active means to resolve traversability failures or clear paths when the intended navigation route is blocked. 

\subsection{Interactive Navigation}
More recently, there has been growing interest in interactive navigation, which moves beyond avoidance-only navigation by allowing robots to deliberately manipulate obstacles to reach a goal~\cite{stilman2005navigation}. Early efforts in learning-based interactive navigation, formalized by works such as the iGibson 0.5 benchmark \cite{xia2020igibson05}, primarily utilized reinforcement learning (RL) and hierarchical methods like HRL4IN and others \cite{li2019hrl4inhierarchicalreinforcementlearning, wang2024interactive} to learn interaction policies like pushing obstacles, pressing buttons, grasping obstructing objects, or opening doors. However, these works have a high dependency on large-scale training in simulation and task-specific reward design, and do not explicitly demonstrate real-world deployment. Follow-up works like~\citet{zhang2024interactive} use VLMs to reason over a scene and construct costmaps; however, this approach oversimplifies objects as being either traversable or non-traversable, which rules out active manipulation. IN-Sight \cite{schoch2024insightinteractivenavigationsight} similarly uses self-supervised training in simulation to learn object-oriented traversability scores, and then does sim2real transfer and integration with classical planning pipelines for interactive navigation through clutter in the real world. However, this approach is dependent on simulation data and the sim2real gap hampers fine-grained physical understanding. SPIN is another sim2real approach \cite{uppal2024spin} which learns RL policies for manipulation and navigation that are deployed on a Hello Robot mobile manipulator. SPIN specifically learns one policy for controlling its egocentric camera gaze to plan collision-free whole-body motions, along with separate manipulation policies, but critically these are not used to resolve navigation blockages. To address this, approaches such as NAMO-LLM and CoINS \cite{zhang2025namollmefficientnavigationmovable, zhou2026coinscounterfactualinteractivenavigation} have integrated VLMs for high-level reasoning over how to manipulate objects obstructing navigation. While promising, these methods typically abstract objects into fixed, broad semantic categories, discarding context-aware details like fragility, geometry, or portability that are critical for safe manipulation. In contrast, \PlannerName{} enables zero-shot, fine-grained reasoning over a diverse set of interaction primitives by combining VLM-based reasoning with classical planning. This allows for reliable open-world interactive navigation in complex, unseen object arrangements, without \emph{any} training or simulation.

\label{app:hardware}
\begin{figure}
    \centering
    \includegraphics[width=0.6\columnwidth]{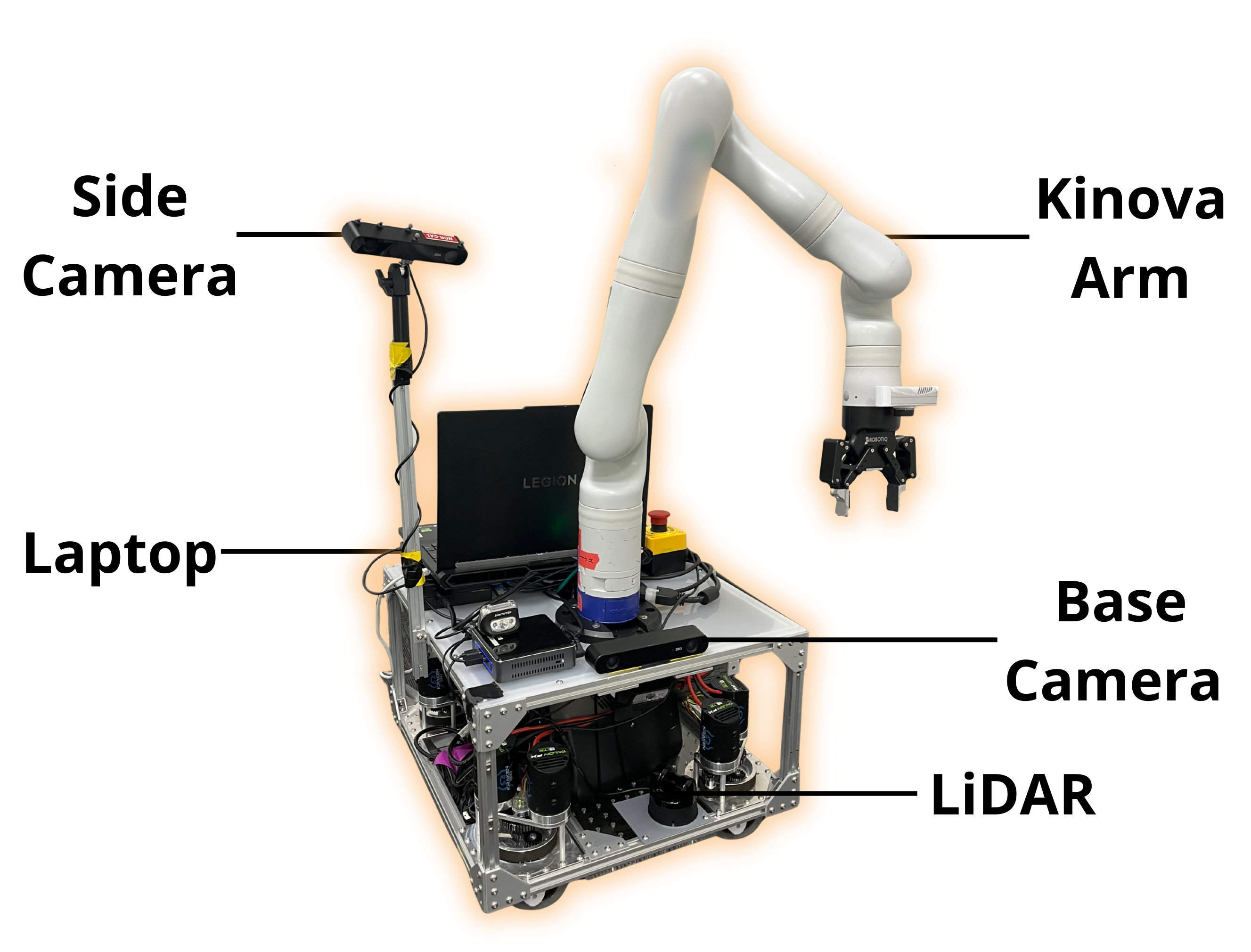}
    \caption{
    \textbf{Hardware:} We deploy on the TidyBot++ holonomic mobile manipulator~\cite{wu2024tidybot++}~with two cameras and a LiDAR for perception. An onboard laptop runs the autonomy stack.}
    \label{fig:hardware}
\end{figure}

\section{Real-World Implementation Details}
\label{appendix:implementation_details}
\subsection{Hardware}
Figure~\ref{fig:hardware} shows the TidyBot++ mobile manipulator \cite{wu2024tidybot++}, the robot used in the paper.

\subsection{Affordance Reasoning} We query \texttt{gemini\allowbreak-3\allowbreak-flash\allowbreak-preview}~\cite{gemini3flash2025} with \texttt{thinking\allowbreak=\allowbreak minimal} using the front-facing camera image. The prompt encodes the robot's physical constraints and evaluates each object through two questions: (1) \emph{can} the robot move it, and (2) \emph{should} it be moved. The full prompt is below.
\begin{GrayVerbatim}
You are analyzing a scene for a mobile robot. The robot needs to understand object properties so it can manipulate the objects in order to navigate effectively.


TASK CONTEXT: {task_description}

Identify all the objects in the scene and assign labels (interactive and non-interactive) based on the following criteria. Ignore the floor and walls.

An object is labeled as interactive if and only if you deem the answer is YES to both questions.
    1. Can the robot physically move the object? Reason about the physical limitations of the robot.
        - The robot is a mobile manipulator with one arm and a 2 fingered gripper. 
        - It can only carry 3kg in the gripper at a time.
        - It can swipe objects with its arm.
If the answer is YES, then verify the second question:
    2. Knowing the robot can move the object, should the robot be moving this object? Answer NO if any of the following are true:
        - Moving the object could damage the robot or the environment
        - The object is fragile, breakable, unstable, damaged, or carefully arranged
        - The object could become damaged if moved
        - The object is expensive or hard to replace
        - The object contents could spill or create a mess if moved
If the answer to any of the questions is deemed to be NO, then the correct label is non-interactive. 

Analyze all the objects that might affect robot navigation or task completion individually. Identify any cases where two objects function together (such as a cup on a table). In these cases, treat them as a single object, create a combined name, and assign one appropriate label.

GROUNDING EXAMPLES:
- Empty wheeled chair → swipe_right(chair)
- Stuffed toy → pick_up_entity(stuffed_toy)
- heavy desk → avoid

Respond with ONLY a JSON object in this exact format:
{{
    "object_properties": {{
        "object_name_1": {{
            "mobility": "interactive_or_non-interactive",
            "interaction": "safe_or_fragile_or_immovable"
        }},
        "object_name_2": {{
            "mobility": "interactive_or_non-interactive",
            "interaction": "safe_or_fragile_or_immovable"
        }}
    }}
}}

Use clear, descriptive names (e.g., "red_box", "glass_vase", "office_chair", "concrete_wall").
\end{GrayVerbatim}

\subsection{Task Planning}
We query \texttt{gemini\allowbreak-3\allowbreak-flash\allowbreak-preview}~\cite{gemini3flash2025} with \texttt{thinking=minimal} using the affordance labels from the affordance reasoning stage and the front-facing camera image to generate a task plan. Note that \texttt{move\_to\_nav2} and \texttt{pick\_up\_entity} in the prompt correspond to \texttt{move\_to} and \texttt{pick\_up} in the main text. The full prompt is below.

\begin{GrayVerbatim}
You are a robot task planner for a mobile robot with one arm and a 2-fingered gripper. The robot constraints:
- It can only carry about 3kg in your gripper at a time
- The gripper can open up to 8.5 cm wide
- The footprint is a 0.8m x 0.8m square

Analyze the image and use the detected object properties to plan your actions. LOOK AT THE IMAGE CAREFULLY to assess FREE SPACE and use this information to inform your planning.
The image is the front-facing view from the robot's perspective.

TASK: {task_description}

DETECTED OBJECT PROPERTIES (from Stage 1):
```json
{properties_str}
```

Based on these properties, decide your interaction strategy:
- INTERACTIVE + SAFE: Can pick or swipe 
- NON-INTERACTIVE + FRAGILE/IMMOVABLE: Must avoid and maintain safe distance

For INTERACTIVE objects, choose action based on efficiency and context:
- Pick: Objects that can be easily grasped and carried by the robot
- Swipe: Larger interactive objects - quick lateral clearing motion. 
Note: You can NOT swipe objects below 0.3 meters in size.

AVAILABLE MOTION PRIMITIVES:
- move_to_nav2(x, y): Move to position (x, y) in meters
- move_to_nav2(object_name, dx, dy): Move to object with offset (dx, dy in meters)
- pick_up_entity(object_name): Pick up the named object
- swipe_left(object_name): Swipe object to the left
- swipe_right(object_name): Swipe object to the right

NAVIGATION OFFSETS:
- Robot Origin: The robot starts at (0.0, 0.0) facing in the +x direction
- Coordinate System:
    - +x: Forward movement (in front of the robot)
    - -x: Backward movement (behind the robot)
    - +y: Leftward movement (to the robot's left)
    - -y: Rightward movement (to the robot's right)
- dx: Forward/Backward offset in meters (positive = forward)
- dy: Left/Right offset in meters (positive = left)
- Want to be approximately 0.6m away from objects for manipulation
- Example: move_to_nav2(obstacle, -0.6, 0.0) → position 0.6m distance from the obstacle (approaching it, stopping short)

PLANNING GUIDELINES:
- For movable objects, YOU decide pick vs swipe based on apparent size/weight
- Position ~0.6m away from objects for manipulation
- Respect fragile/hazardous classifications from Stage 1
- You may not need to explicitly interact with all interactive objects - only those that necessarily blocking the path you want to take.
- move_to_nav2 primitive calls a navigation planner that avoids obstacles
- IMPORTANT: move_to_nav2 finds navigation sub goals so there is DO NOT give intermediate waypoints

GROUNDING EXAMPLES:
- Empty wheeled chair → swipe_right(chair)
- Stuffed toy → pick_up_entity(stuffed_toy)
- heavy desk → avoid

Respond with ONLY a JSON object in this exact format:
{{
    "reasoning": "Explain your planning strategy based on object properties and visual analysis",
    "geometric_analysis": "Analyze spatial constraints and navigation paths",
    "interaction_modes": {{
        "object_name": "pick_or_swipe_or_avoid"
    }},
    "property_overrides": ["List any property assessments you're overriding and why"],
    "plan": [
        "move_to_nav2(x, y)",
        "action_3"
    ]
}}
\end{GrayVerbatim}

\subsection{Open-Vocabulary Keypoint Detection and Segmentation}

We use \texttt{gemini-\allowbreak2.5-\allowbreak pro}~\cite{gemini25pro2025} for both 2D keypoint detection and segmentation to condition the motion primitives. As these queries generally occur when the robot is positioned close to the target object for manipulation, we first query the side arm-mounted camera, which has better visibility of the area directly in front of the robot. If the object is not found, we fall back to the front-facing base camera. 

The prompt for 2D keypoint detection is as follows: 
\begin{figure}
    \centering    \includegraphics[width=0.8\columnwidth]{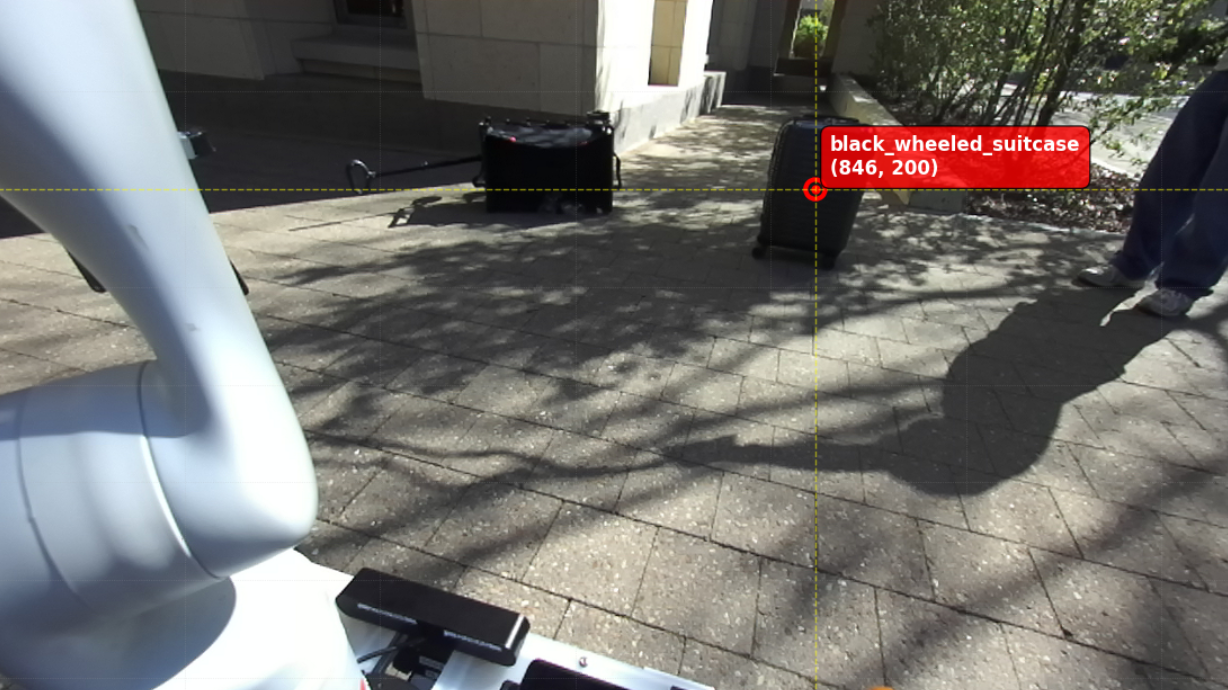}
    \caption{\textbf{Example Open-Vocabulary Keypoint Detection.} We use \texttt{gemini-\allowbreak2.5-\allowbreak pro}~\cite{gemini25pro2025} to detect keypoints of objects specified by the task planner through language queries.}
    \label{fig:gemini_pointing}
\end{figure}
\begin{GrayVerbatim}
Point to the {prompt}. The {prompt} may have gaps or be partially occluded.
Make sure your point is on a visible part of the {prompt}.
The answer should follow the json format: "
'[{"point": <point>, "label": <label1>}, ...]. '
"The points are in [y, x] format normalized to 0-1000.
\end{GrayVerbatim}
Figure \ref{fig:gemini_pointing} shows an example result of 2D keypoint detection on a black wheeled suitcase.

The prompt for 2D segmentation is as follows:
\begin{GrayVerbatim}
Find the following objects in this image: {objects_str}. For each
object, give me ONLY ONE segmentation mask for the single instance
you are most confident about. Focus ONLY on the foreground objects
and exclude any background elements like walls, floors, or surfaces.
The masks should tightly bound just the objects themselves, not any
background or supporting surfaces. Output a JSON list where each
entry contains the 2D bounding box in the key "box_2d", the
segmentation mask in key "mask", and the text label in the key
"label". Choose the clearest, most well-defined instance of each
object type and ignore any other instances. IMPORTANT: Use the EXACT
labels I provided in quotes - do not modify them. For example, if I
ask for "red block", your label must be exactly "red block", not
"red cube" or anything else.
\end{GrayVerbatim}

\subsection{Motion Primitives}
In this instantiation of the framework, we design a small set of motion primitives that can address a wide range of interactive navigation scenarios:
\begin{itemize}
    \item \texttt{move\_to()}: Navigate to a world-frame coordinate, or offset from an object. For object-relative navigation, we use the VLM's pointing module to localize the object in 3D and compute the offset. We use the ROS2 Nav2 Navigation Stack \cite{macenski2020marathon2} with a LiDAR sensor for obstacle avoidance. 
    \item \texttt{pick\_up()}: Grasp and carry an object with the robot arm. We use the VLM's pointing module for the grasping point and the VLM's segmentation to compute the object's 3D point cloud and principal axes for grasp orientation. 
    \item \texttt{swipe\_left() / swipe\_right()}: Clear an object with the arm in a lateral direction. We use the VLM's segmentation to obtain the object's 3D point cloud to compute the approach position for swiping, a heuristic distance away from the object.
\end{itemize}

\subsection{LiDAR-based Navigation Stack}
We provide details on the navigation stack used both as the \texttt{height-only} baseline and within the \texttt{move\_to} primitive. We use a Unitree L2 LiDAR for odometry estimation via FAST-LIO~\cite{xu2020fastlio}. The registered LiDAR points are projected to 2D and aggregated into an occupancy grid, which is used to compute the costmap. Table~\ref{tab:nav_params} lists the relevant parameters. We use the ROS2 Nav2 Navigation Stack~\cite{macenski2020marathon2} for path planning and follow the resulting plan with a holonomic controller without rotation, maintaining consistent camera views of objects identified during planning.
\begin{table}[]
\centering
\caption{\textbf{Navigation Costmap Parameters.}}
\label{tab:nav_params}
\small
\begin{tabular}{l l}
\toprule
\textbf{Parameter} & \textbf{Value} \\
\midrule
Map resolution & 0.05 m/pixel \\
Map size & 30m $\times$ 30m \\
Max laser range & 10m \\
\bottomrule
\end{tabular}
\end{table}

\section{Affordance Reasoning Experiment Details}
\label{appendix:affordance-reasoning-expt}
\subsection{Evaluation Set} 
\begin{figure}
    \centering
    \includegraphics[width=\columnwidth]{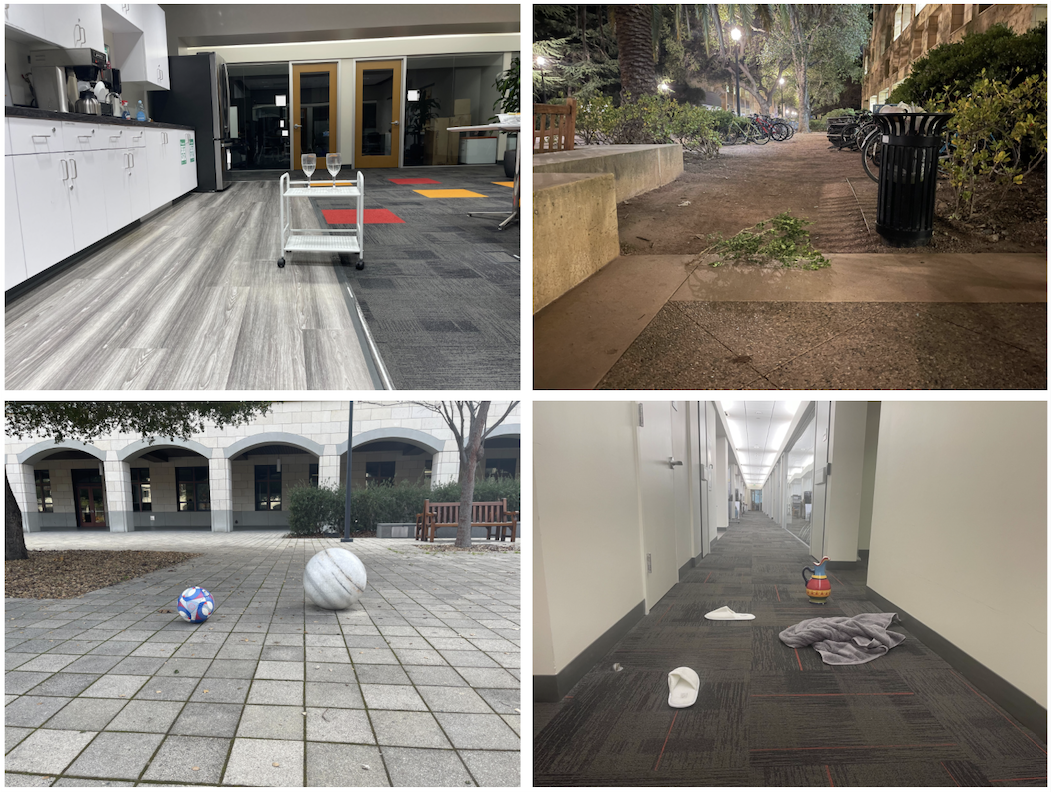}
    \caption{\textbf{Sample Images from the Affordance Reasoning Evaluation Set.} To evaluate open-world interactive navigation, we collect images spanning three types of diversity: objects, environments, and contextual variations. 
    }
    \label{fig:affordance_sample}
\end{figure}

To test affordance reasoning for \textit{open-world} interactive navigation, we collect a dataset of 100 images covering a range of objects, environments, and contextual variations. Figure \ref{fig:affordance_sample} shows sample images in the evaluation set. Table~\ref{tab:benchmark_composition} summarizes the composition of the evaluation set. All affordance labels
were annotated by an expert annotator familiar with the robot's physical capabilities. Table \ref{tab:context_variations} lists examples of how the context variations of the same object type in the evaluation set affect its affordance. For example, a default rolling cart is interactive, but one with a ``DO NOT TOUCH'' sign is not.

\begin{table}[t]
\centering
\caption{\textbf{Composition of the Affordance Reasoning Benchmark.} Our benchmark spans diverse environments, lighting, clutter, and obstacle sizes.}
\label{tab:benchmark_composition}
\small
\begin{tabular}{l l}
\toprule
\textbf{Attribute} & \textbf{Composition} \\
\midrule
Scene type & Indoor (69\%); outdoor (31\%) \\
\midrule
Lighting & Bright artificial (66\%); natural (28\%); \\
 & dim (6\%) \\
\midrule
Object affordance & \orangeI{} 94 / 217 (43.3\%); \\
 & \grayN{} 123 / 217 (56.7\%) \\
\midrule
Clutter & 1--4 foreground objects per scene \\
\midrule
Obstacle size & Small electronics to large trees \\
\bottomrule
\end{tabular}
\end{table}

\begin{table}[t]
\centering
\caption{\textbf{Example Context Variations in Affordance Reasoning Evaluation Set.} The same object type is marked as \textcolor{orange}{\textbf{interactive (I)}} or \textcolor{gray}{\textbf{non-interactive (N)}} depending on context.}
\label{tab:context_variations}
\small
\begin{tabular}{l l}
\toprule
\textbf{Object} & \textbf{Contexts} \\
\midrule
Rolling cart & Default \orangeI; w/ glassware \grayN; \\
 & w/ glasses \grayN; w/ stuffed toys \orangeI; \\
 & w/ toys + ``DO NOT TOUCH'' sign \grayN; \\
 & w/ heavy robot arm \grayN; \\ & w/ snack boxes \orangeI\\
\midrule
Wheeled crate & Default \orangeI; w/ toaster box precariously on edge \grayN; \\
 & w/ bundled wire on top \orangeI; \\
 & w/ cable attached \grayN \\
\midrule
Suitcase & Default \orangeI; w/ caution tape \grayN \\
\midrule
Chair & Default \orangeI; w/ glass vase on seat \grayN \\
\midrule
Spheres & Soccer ball \orangeI; marble sphere \grayN \\
\bottomrule
\end{tabular}
\end{table}

\subsection{Details for the \texttt{closed-set} Baseline \cite{zhang2024interactive}}
The original approach in \cite{zhang2024interactive} relies on per-scene human instructions (e.g., ``please go through the curtain and be careful of the chair."). As per-scene instructions are unavailable in our zero-shot evaluation setting, we instead generate a fixed list of 100 object categories (50 interactive, 50 non-interactive) using \texttt{gemini-3-flash-preview} ~\cite{gemini3flash2025} with \PlannerName{}'s affordance reasoning prompt as context. This ensures the closed-set categories reflect the same affordance reasoning principles as our method, while still producing a fixed category-level list independent of scene-specific instructions. We also replace their traversable/non-traversable labels with interactive/non-interactive to match our task formulation.

The prompt is as follows:

\begin{GrayVerbatim}
Based on this context which has affordance instructions and grounding examples: '{affordance reasoning prompt}', please make an exhaustive prompt of 50 categories each based on common objects you may encounter on the floor. (50 objects that you can interact with, 50 objects you should avoid). Format it as a collection of sentences. For example - Interact with items like empty wheeled chair, stuffed toy, .... Avoid or be careful of items like heavy desk, ....
\end{GrayVerbatim}

This results in the following prompt which is used by the \texttt{closed-set} baseline: 
\begin{GrayVerbatim}
Interact with items like an empty wheeled chair, a stuffed toy, a small cardboard box, a plastic water bottle, a lightweight laundry basket, a tennis ball, a pair of sneakers, a discarded newspaper, a plastic food container, a small throw pillow, a light wooden stool, a yoga block, a slippers, a handheld remote control, a small plastic trash bin, a folded towel, a soft-sided cooler bag, a frisbee, a light plastic bucket, a rolling office footrest, a small backpack, a notebook, a plastic toy truck, a beanbag chair, a light yoga mat, a cardboard mailing tube, a small area rug, a plastic watering can, a light step stool, a handheld umbrella, a soft lunchbox, a plastic storage bin lid, a small pet bed, a light wooden crate, a foam roller, a plastic dustpan, a small dry mop, a lightweight folding chair, a soft briefcase, a plastic tray, a small cardboard tray, a light plastic stool, a rolled-up poster, a small fabric ottoman, a plastic toy block, a light wicker basket, a small duffel bag, a plastic clipboard, a light broom, and a small empty flower pot. 
Avoid or be careful of items like a heavy desk, a glass vase, a ceramic floor lamp, a large bookshelf, a heavy filing cabinet, a television set, a desktop computer tower, a large potted plant, a glass coffee table, a heavy wooden dining table, a refrigerator, a microwave, a space heater, a floor mirror, a fragile porcelain figurine, a stack of heavy textbooks, a metal safe, a heavy toolbox, a piano, a large aquarium, a marble pedestal, a heavy leather sofa, a treadmill, a stationary bike, a large chest of drawers, a fragile glass sculpture, a heavy radiator, a wall-mounted heater, a large speaker, a heavy metal trunk, a delicate antique chair, a floor-standing clock, a heavy printer, a wine rack with bottles, a fragile drying rack, a heavy workbench, a large trash compactor, a heavy stone statue, a fragile folding screen, a heavy air conditioner unit, a large dog crate, a heavy weight bench, a stack of ceramic tiles, a large bag of concrete, a heavy industrial fan, a fragile neon sign, a heavy metal shelving unit, a large wooden wardrobe, a fragile model ship, and a heavy cast-iron stove.

\end{GrayVerbatim}

\subsection{Details for the \texttt{closed-set-oracle} Baseline \cite{zhang2024interactive}}
This represents the upper bound of the above baseline. It uses a manually curated list containing all object categories present in our evaluation set with their ground-truth category-level interaction labels. While no objects are missed, affordances remain fixed per category and do not account for context-dependent variations. 
The prompt for the affordance reasoning benchmark is as follows:
\begin{GrayVerbatim}
Interact with items like a suitcase, wheeled crate, soccer ball, stuffed toy, vegetables, fruits, slippers, shoes, towel, napkin, trash bag, clothing, small table, bundled jump rope, plastic stool, tongs, microwave, pencil sharpener, waste basket, paper, small basket, small chair, rolling chair, rolling cart, snack box, cardboard box, trash bin, ball, beanie, wheeled chair, slipper, branch, tennis balls, empty bag, plastic cups, markers, binders, car toy, pillow, wheeled table, jeans, foldable stool, short wire, bag, toy blocks, toy trains, small stool. 

Avoid or be careful of items like a laptop, large table, vase, ceramic dishes, glass item, electrical appliance, bundle of wires, scooter, bike, bush, bag of leaves, stone statue, heavy dumbbell, pole, newspaper stand, shovel, cinder block, laptop box, plants, robot arm, unfurled wired cables, sofa, tree, bench, human, shop vacuum, picnic table, electronics.

\end{GrayVerbatim}

\begin{table}[]
\centering
\caption{\textbf{Affordance Reasoning Comparison}. Number of objects with correctly classified interactivity. \textbf{Simple} cases contain objects in their default state, while \textbf{Complex} cases involve context-dependent configurations (e.g., cart with fragile items, table with vase, suitcase with caution tape).}
\label{tab:object_complexity}
\small
\begin{tabular}{lcc!{\color{gray}\vrule width 0.4pt}c}
\toprule
 & \textbf{Simple} & \textbf{Complex} & \textbf{Total} \\
\midrule
\texttt{closed-set}          & 30 / 112  & 23 / 105 & 53 / 217  \\
\texttt{closed-set-oracle}   & 106 / 112 & 38 / 105 & 144 / 217 \\
\texttt{height-only}         & 57 / 112  & 73 / 105 & 130 / 217 \\
\texttt{\textbf{MessyNav (ours)}} & \textbf{101 / 112} & \textbf{85 / 105} & \textbf{186 / 217} \\
\bottomrule
\end{tabular}
\end{table}

\begin{table*}[t!]
\centering
\caption{\textbf{Real-World Navigation Scenarios.} To evaluate open-world interactive navigation, we test in a bedroom and conference room environments while deliberately varying objects, contexts, and required interactions across trials. The scenes also contain furniture (e.g., couch, bed, drawers, meeting tables) as fixed non-interactive obstacles. Additional objects are listed and marked as \textcolor{orange}{\textbf{interactive (I)}} or \textcolor{gray}{\textbf{non-interactive (N)}}.}
\label{tab:nav_scenarios}
\small
\begin{tabular}{l p{7.5cm} p{3cm}}
\toprule
\textbf{Environment} & \textbf{Additional Objects in Scenario} & \textbf{Interaction Needed} \\
\midrule
{Messy Bedroom} & Cart w/ broken wheels \grayN, Cart \orangeI  & Swipe cart \\
 & Computer Box \grayN, Clear Box \orangeI  & Swipe clear box \\
  & Suitcase balancing books \grayN, Cart w/ one book \orangeI  & Swipe cart \\
  & Crate w/ one book \orangeI, Cart w/ loose golf balls \grayN  & Swipe crate\\
  & Small full box \grayN, Large clear box \orangeI  & Swipe clear box \\
  & Wheeled drawer w/ one toy block \orangeI, Cart w/ toy castle \grayN  & Swipe wheeled drawer\\
  & Small clear box w/ vase \grayN, Clear box w/ plush toy \orangeI  & Swipe clear box \\
  & Cart tangled in wire \grayN, Crate w/ stuffed toy \orangeI  & Swipe crate \\
  & Cart w/ clear box holding items \orangeI, Crate w/ toy castle \grayN  & Swipe cart w/ clear box\\
  & Cart w/ active fan \grayN, Cart w/ bundled wire \orangeI  & Swipe cart w/ wire\\
\midrule
{Conference Room} & Cart w/ laptop \grayN, Crate w/ pitcher in clear box \orangeI  & Swipe crate \\
 & Cart w/ items in box \orangeI, Cart w/ glass of water \grayN  & Swipe cart w/ box \\
  & Suitcase w/ caution tape \grayN, Suitcase \orangeI  & Swipe suitcase\\
  & Suitcase w/ bag \orangeI, Cart w/ loose water bottles in bag \grayN  & Swipe suitcase\\
  & Suitcase w/ caution tape \grayN  & None\\
\bottomrule
\end{tabular}
\end{table*}

\subsection{Additional Result Analysis}
Table \ref{tab:object_complexity} breaks down affordance reasoning accuracy into Simple and Complex cases. Simple cases are objects in their default state (e.g., an empty cart, a chair with nothing on it), while Complex cases involve context-dependent configurations (e.g., a cart with electronics precariously placed on top, a suitcase with caution tape on top). \PlannerName{} is competitive on simple cases and substantially outperforms all baselines on complex cases. The results highlight the gap between \PlannerName{} and the strongest baseline, \texttt{closed-set-oracle}. Despite having access to privileged information about object categories in the evaluation set, \texttt{closed-set-oracle} achieves comparable performance only on simple cases. Its performance degrades significantly on complex scenes as it is unable to reason beyond its default context. In contrast, \PlannerName{} requires no predefined object categories and can reason about the object's broader context.

\section{Details on Real-World Comparative Experiments}
\subsection{Setup} The navigation goal is specified as a world-frame coordinate in the task instruction: \texttt{"Navigate to (goal\_x, goal\_y) and handle any obstacles in your path."} We mark environment boundaries beyond the navigable area as occupied in the occupancy map. To evaluate open-world interactive navigation, we test across 15 scenarios in two environments that vary in objects, contexts, and required interactions. For example, the same cart is empty in one scenario and holds a glass of water in another. Table~\ref{tab:nav_scenarios} lists the full set of scenarios.

\subsection{Details for the \texttt{closed-set-oracle} Baseline \cite{zhang2024interactive}}
The prompt for real-world comparative experiments is as follows:
\begin{GrayVerbatim}
Interact with items like a cart, stuffed toy, storage bin, cardboard box, toys, stuffed toys, toy blocks, suitcase, plastic bin, tissue, paper bags, books, golf ball, tennis ball, tennis ball canister, plate, wheeled crate, clothes, stool, chair, water bottles, bottles, and bowls. Avoid or be careful of items like wires, electrical appliances, drawer, bed, laptop, glass items, pitcher, heavy furniture and vase.
\end{GrayVerbatim}
Subsequently, the results are passed to the VLM planner used in \PlannerName{}.

\subsection{Details for the \texttt{vlm-planner-only} Baseline}
The prompt is as follows:
\begin{GrayVerbatim}
You are a robot task planner for a mobile robot with one arm and a 2-fingered gripper. The robot constraints:
- It can only carry about 3kg in your gripper at a time
- The gripper can open up to 8.5 cm wide
- The footprint is a 0.8m x 0.8m square

Analyze the image and plan your actions. LOOK AT THE IMAGE CAREFULLY to assess FREE SPACE and use this information to inform your planning.
The image is the front-facing view from the robot's perspective.

TASK: {task_description}

AVAILABLE MOTION PRIMITIVES:
- move_to_nav2(x, y): Move to position (x, y) in meters
- move_to_nav2(object_name, dx, dy): Move to object with offset (dx, dy in meters)
- pick_up_entity(object_name): Pick up the named object
- swipe_left(object_name): Swipe object to the left
- swipe_right(object_name): Swipe object to the right
When interacting with objects, choose action based on efficiency and context:
- Pick: Objects that can be easily grasped and carried by the robot
- Swipe: Larger objects - quick lateral clearing motion. 
Note: You can NOT swipe objects below 0.3 meters in size.

NAVIGATION OFFSETS:
- Robot Origin: The robot starts at (0.0, 0.0) facing in the +x direction
- Coordinate System:
    - +x: Forward movement (in front of the robot)
    - -x: Backward movement (behind the robot)
    - +y: Leftward movement (to the robot's left)
    - -y: Rightward movement (to the robot's right)
- dx: Forward/Backward offset in meters (positive = forward)
- dy: Left/Right offset in meters (positive = left)
- Want to be approximately 0.6m away from objects for manipulation
- Example: move_to_nav2(obstacle, -0.6, 0.0) -> position 0.6m distance from the obstacle (approaching it, stopping short)

PLANNING GUIDELINES:
- For movable objects, YOU decide pick vs swipe based on apparent size/weight
- Position ~0.6m away from objects for manipulation
- You may not need to explicitly interact with all objects - only those that necessarily blocking the path you want to take.
- move_to_nav2 primitive calls a navigation planner that avoids obstacles
- IMPORTANT: move_to_nav2 finds navigation sub goals so there is DO NOT give intermediate waypoints

GROUNDING EXAMPLES:
- Empty wheeled chair -> swipe_right(chair)
- Stuffed toy -> pick_up_entity(stuffed_toy)
- heavy desk -> avoid

Respond with ONLY a JSON object in this exact format:
{{
"reasoning": "Explain your planning strategy based on visual analysis",
"geometric_analysis": "Analyze spatial constraints and navigation paths",
"plan": [
    "move_to_nav2(x, y)",
    "action_3"
]
}}

\end{GrayVerbatim}

\begin{figure}[]
    \centering
    \includegraphics[width=\linewidth]{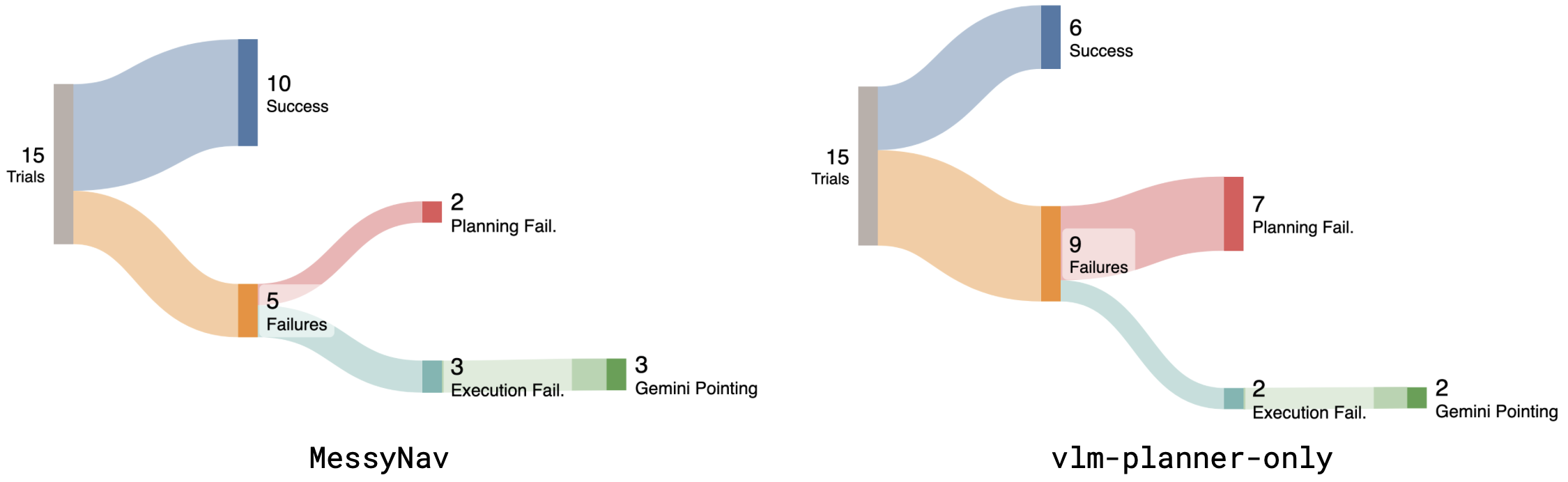}
    \caption{\textbf{\PlannerName{}'s failure breakdown compared to the strongest baseline.} Sankey diagrams of trial outcomes for \PlannerName{} (left) and \texttt{vlm-planner-only} (right) over 15 real-world trials. \PlannerName{} succeeds on 10/15 trials versus 6/15 for \texttt{vlm-planner-only}. While most of \texttt{vlm-planner-only}'s failures are planning failures in which \texttt{vlm-planner-only} decides to interact with non-interactive objects (7 of 9), \PlannerName{} reduces these to 2, showcasing the importance of explicit affordance reasoning.}
    \label{fig:sankey_trial_outcomes}
\end{figure}
\subsection{Failure Mode Analysis}
Figure~\ref{fig:sankey_trial_outcomes} compares failure modes between \PlannerName{} and the strongest baseline, \texttt{vlm-planner-only}. Overall, \PlannerName{} outperforms \texttt{vlm-planner-only}, particularly in planning success. As \texttt{vlm-planner-only} lacks explicit affordance reasoning, it incurs 7 planning failures rather than 2. Most of \PlannerName{}'s remaining failures instead stem from incorrect Gemini pointing (e.g., pointing to free space within a cart, resulting in failed execution). Its two planning failures arise from subtle affordance-reasoning errors, such as failing to distinguish a cart with a missing small wheel from an intact one.

\section{Details on the Offline Planning Evaluation}
\label{sec:scale-details}

\subsection{Scene Set}
In addition to the 15 scenes used in the real-world experiments (Table~\ref{tab:nav_scenarios}), we include 40 additional scenes from a kitchen, an outdoor daytime setting, and an outdoor nighttime setting, for a total of 55 scenes across five environments. Each scene contains at least one object blocking the direct path to the goal. We vary object contexts so that deciding whether and how to interact requires reasoning beyond object category.

\subsection{Protocol} For each scene, both methods receive the same egocentric RGB observation
and goal specification and generate a high-level plan. A human evaluator
assesses each plan under the assumption of perfect low-level execution.
A plan is successful if it would reach the goal without interacting with
any object labeled non-interactive in the ground truth. This isolates
high-level planning performance from low-level execution failures.
We conduct 15 runs per method, where each run evaluates all 55 scenes once, for a total of 825 scene-level evaluations per method.

\end{document}